\documentclass[sigconf,natbib=false,printacmref=false]{acmart}
\AtBeginDocument{%
  }

\setcopyright{acmlicensed}
\copyrightyear{2024}
\acmYear{2024}
\acmDOI{XXXXXXX.XXXXXXX}

\RequirePackage[
  datamodel=acmdatamodel,
  style=acmnumeric,
  giveninits=true
  ]{biblatex}

\usepackage{booktabs}   
\usepackage{bm}
\usepackage{subfigure}
\usepackage{graphicx}
\usepackage{algorithm}
\usepackage{algorithmic}
\usepackage{multirow}
\usepackage{upgreek}
\usepackage{bm}

\begin{document}

\title{Escaping Local Optima in Global Placement}


\author{Ke Xue}
\authornote{Equal contribution}
\email{xuek@lamda.nju.edu.cn}
\affiliation{
  \institution{National Key Laboratory for Novel Software Technology, Nanjing University}
  \institution{School of Artificial Intelligence, Nanjing University}
  \country{China}
}

\author{Xi Lin}
\authornotemark[1]
\email{211300083@smail.nju.edu.cn}
\affiliation{
  \institution{National Key Laboratory for Novel Software Technology, Nanjing University}
  \institution{School of Artificial Intelligence, Nanjing University}
  \country{China}
}

\author{Yunqi Shi}
\email{shiyq@lamda.nju.edu.cn}
\affiliation{
  \institution{National Key Laboratory for Novel Software Technology, Nanjing University}
  \institution{School of Artificial Intelligence, Nanjing University}
  \country{China}
}

\author{Shixiong Kai}
\email{kaishixiong@huawei.com}
\affiliation{
  \institution{Huawei Noah’s Ark Lab}
  \country{China}
}

\author{Siyuan Xu}
\email{xusiyuan520@huawei.com}
\affiliation{
  \institution{Huawei Noah’s Ark Lab}
  \country{China}
}

\author{Chao Qian}
\authornote{Corresponding author}
\email{qianc@lamda.nju.edu.cn}
\affiliation{
  \institution{National Key Laboratory for Novel Software Technology, Nanjing University}
  \institution{School of Artificial Intelligence, Nanjing University}
  \country{China}
}


% \renewcommand{\shortauthors}{Anonymous}

\begin{abstract}
Placement is crucial in the physical design, as it greatly affects power, performance, and area metrics. Recent advancements in analytical methods, such as DREAMPlace, have demonstrated impressive performance in global placement. However, DREAMPlace has some limitations, e.g., may not guarantee legalizable placements under the same settings, leading to fragile and unpredictable results. This paper highlights the main issue as being stuck in local optima, and proposes a hybrid optimization framework to efficiently escape the local optima, by perturbing the placement result iteratively. The proposed framework achieves significant improvements compared to state-of-the-art methods on two popular benchmarks.
\end{abstract}

\keywords{Placement, Physical Design, Non-convex Optimization, Hybrid Optimization}

\maketitle

\section{Introduction}

Placement is crucial in the physical design of very large-scale integration (VLSI) circuits. It decides the physical locations of standard cells in the layout, which will significantly affect the solution space of the succeeding routing stages, thus greatly affecting power, performance, and area (PPA) metrics. Placement usually consists of three stages: 1) global placement provides rough locations of standard cells with the aim of minimizing wirelength; 2) legalization then removes overlaps and design rule violations based on the global placement solution; 3) detailed placement finally incrementally improves the solution quality. Among these stages, global placement provides a fundamental solution for the subsequent processes, thus plays an important role.

Among different types of global placement methods, analytical global placers have been investigated for decades and are used as the default global placer due to their superior performance. One mainstream analytical placers are quadratic placers, which use a quadratic function to represent wirelength and mitigate cell density~\cite{ripple,polar}. 
Although quadratic placers converge quickly, the quality of their solutions is limited due to the low modeling order of wirelength. Non-linear placers~\cite{chen2008ntuplace3,lu2015eplace,cheng2018replace} use a smooth variant of the halfperimeter wirelength (HPWL) metric to approximate wirelength, produce higher solution quality but have a large runtime overhead.
With the rapid development of GPU devices, acceleration by GPU becomes an important direction. Recently, DREAMPlace~\cite{lin2020dreamplace,dreamplace4} accelerated ePlace/RePlace~\cite{lu2015eplace,cheng2018replace} on GPU by casting the placement problem as a neural network training problem and demonstrated the superiority of GPU accelerated global placers.

\begin{figure}[t!]
\centering
\includegraphics[width=0.49\textwidth]{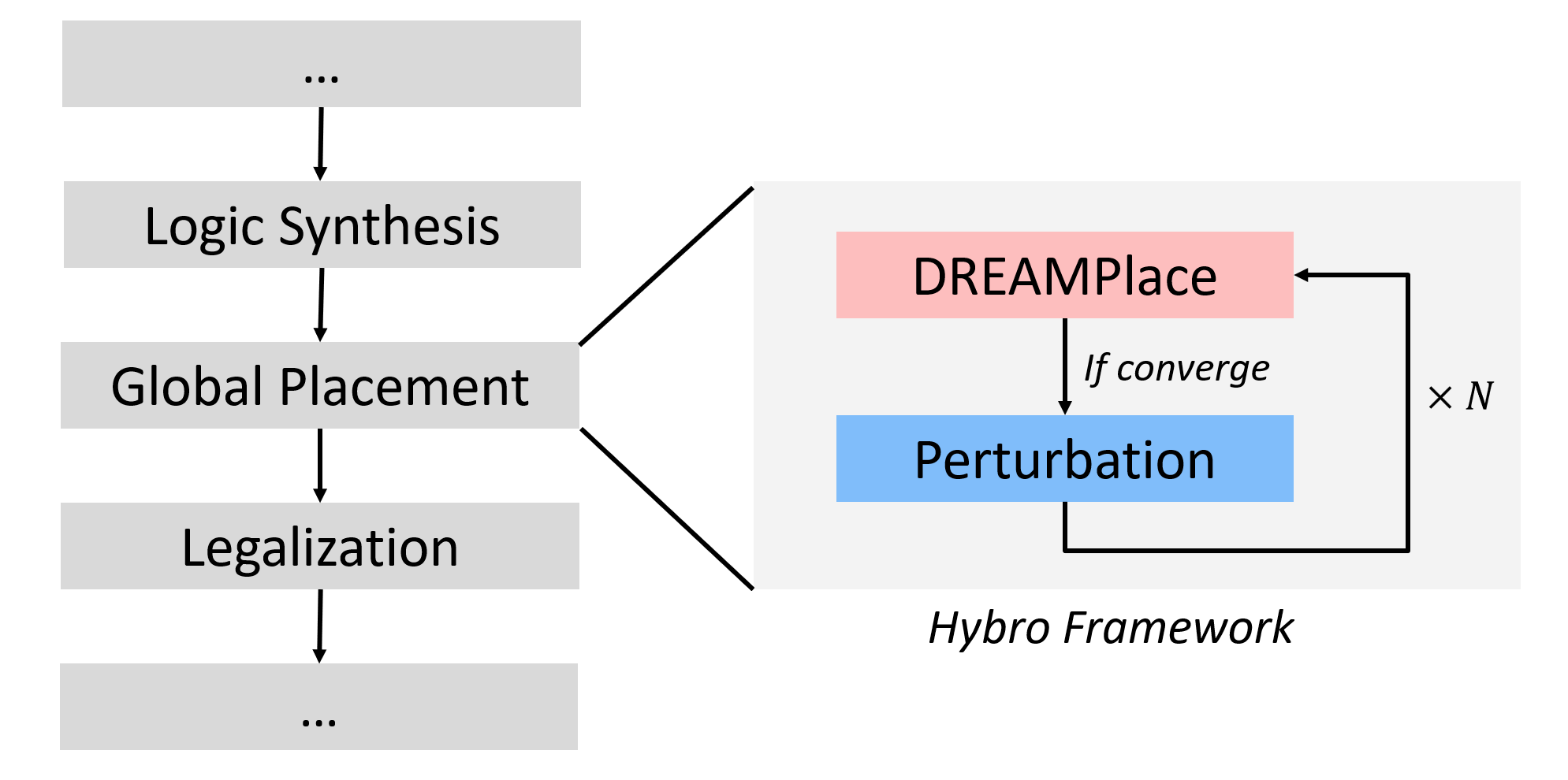}
\caption{Hybro-powered design flow, which can help in escaping local optima in global placement. If the gradient-based global placer (e.g., DREAMPlace) converges, Hybro adds perturbation to the placement result. This procedure will be repeated for $N$ iterations.}
% todo change the font size of the figure
\label{fig:hybro_framework}
\end{figure}

However, DREAMPlace still has some limitations. For example, when using the same settings, the random seeds have a significant impact on DREAMPlace's performance and may not even guarantee legalizable placements, leading to fragile and unpredictable results~\cite{lai2022maskplace,agnesina2023autodmp}. In this paper, we highlight the main issue for the unsatisfactory performance of DREAMPlace as being stuck in local optima, which is a common issue when using gradient-based optimizer to solve non-convex optimization problems~\cite{on_nonconvex}. To address this issue, we propose a hybrid optimization framework (called Hybro) to efficiently escape the local optima, by perturbing the placement result iteratively, as shown in Figure~\ref{fig:hybro_framework}. 

The major contributions are summarized as follows.
\begin{itemize}
    \item We point out that one of the main issues of analytical methods, e.g., DREAMPlace, is stuck into local optima in the global placement stage. 
    \item To address this issue, we propose a Hybrid Optimization framework for global placement (Hybro) and several corresponding perturbation strategies (i.e., Hybro-Shuffle and Hybro-WireMask) to help in escaping local optima.
    \item Experiments on two popular benchmarks, i.e., ISPD 2005~\cite{nam2005ispd2005} and ICCAD 2015~\cite{iccad15}, demonstrate the effectiveness of Hybro framework, where Hybro-WireMask not only has the better full HPWL value, but also achieves better timing and congestion metrics. 
\end{itemize}

\section{Related work}
Analytical methods~\cite{essential-issues-in-analytical} place macros and standard cells simultaneously, which can be roughly categorized into quadratic placement and nonlinear placement. 
Quadratic placement~\cite{ripple,polar} iterates between an unconstrained quadratic programming phase to minimize wirelength and a heuristic spreading phase to remove overlaps. 
Nonlinear placement~\cite{chen2008ntuplace3,lu2015eplace,cheng2018replace} formulates a nnonlinear optimization problem and tries to directly solve it with gradient descent methods. Generally speaking, nonlinear placement can achieve better solution quality, while quadratic placement is more efficient. Recently, there has been extensive attention on GPU-accelerated non-linear placement methods. One notable algorithm is DREAMPlace~\cite{lin2020dreamplace,dreamplace4}, which has demonstrated state-of-the-art performance.

Black-box optimization methods for placement have a long history. Earlier methods such as SP~\cite{murata1996vlsi} and B$^*$-tree~\cite{chang2000b} have poor scalability due to the rectangular packing formulation. Recently, some black-box optimization methods have made significant progress by changing the search space. 
AutoDMP~\cite{agnesina2023autodmp} improves DREAMPlace by using Bayesian optimization to explore the configuration space and shows remarkable performance on multiple benchmarks. 
WireMask-BBO~\cite{wiremask-bbo} is a recently proposed macro placement method, which adopts a wire-mask-guided greedy genotype-phenotype mapping and can be equipped with any BBO algorithm, demonstrating the superior performance over packing-based, reinforcement learning, and analytical methods. 

\section{Preliminaries}
The circuit in the global placement stage is considered as a graph where vertices model gates. The main input information is the netlist $\mathcal {N}=(V,E)$, where $V$ denotes the information (i.e., height and width) about all cells designated for placement on the chip, and $E$ is a hyper-graph comprised of nets $e_i\in E$, which encompasses multiple cells (including both macros and standard cells) and denotes their inter-connectivity in the routing stage. Given a netlist, a fixed canvas layout and a standard cell library, a global placement method is expected to determine the appropriate physical locations of movable cells such that the total wirelength can be minimized. 

A global placement solution $\bm{s}=\{(x_1,y_1), \dots, (x_k, y_k)\}$ consists of the positions of all the cells $\{v_i\}_{i=1}^k$, where $k$ denotes the total number of cells. The objective of global placement is to minimize the total HPWL of all the nets while satisfying the cell density constraint, which is formulated as,
\begin{equation}
\min_{\bm{s}} HPWL(\bm{s}) = \min_{\bm{s}} \sum_{e\in E}HPWL_e(\bm{s}), \ 
\text{s.t.} \ D(\bm{s}) \leq \epsilon,
\end{equation}
where $D$ denotes the density, $\epsilon$ is a threshold, and $HPWL_e$ is the HPWL of net $e$, which is defined as:
$ HPWL_e(\bm{s}) = (\max\nolimits_{v_i\in e} x_i - \min\nolimits_{v_i\in e} x_i) + (\max\nolimits_{v_i\in e} y_i - \min\nolimits_{v_i\in e} y_i)$. 

Nonlinear placement usually reformulates the objective with a smooth version of HPWL to facilitate the calculation of gradient~\cite{hsu2011tsv}:
\begin{equation}\label{eq:obj}
\min_{\bm s} \sum_{e\in E} WL_e(\bm s) + \lambda \cdot D(\bm s), 
\end{equation}
where wirelength $WL_e(\bm s) = WL_e(\bm x) + WL_e(\bm y)$, $WL_e(\bm{x})$ is calculated as
\begin{equation}\label{eq:wle}
WL_e(\bm{x}) = \frac{\sum_{i:v_i\in e} x_i \exp(x_i/\gamma)}{\sum_{i:v_i\in e}\exp(x_i/\gamma)} - \frac{\sum_{i:v_i\in e} x_i \exp(-x_i/\gamma)}{\sum_{i:v_i\in e}\exp(-x_i/\gamma)},
\end{equation}
and $WL_e(\bm y)$ is calculated in a similar way. That is, the weighted average model is used for approximating the wirelength, and the coefficient $\gamma$ controls the accuracy of the approximation. The coefficient $\lambda$ in Eq.~(\ref{eq:obj}) controls the importance of density, which can be seen as a penalty term and is usually set to a small value at the beginning of the placement process.

\subsection{DREAMPlace}
Recently, DREAMPlace~\cite{lin2020dreamplace,dreamplace4} transforms the non-linear placement problem in Eq.~(\ref{eq:obj}) into a neural network training problem, solves it by classical gradient descent and leverages deep learning hardware (i.e., GPU) and software toolkit (e.g. PyTorch), enabling ultra-high parallelism and acceleration. Besides, based on ePlace/RePlace~\cite{lu2015eplace,cheng2018replace} algorithms, DREAMPlace can produce state-of-the-art global placement quality. 

However, DREAMPlace still has some limitations, e.g., it cannot ensure that placements will be legal for designs with numerous macros~\cite{lai2022maskplace,agnesina2023autodmp}. Furthermore, the optimization's convergence and final objective value are significantly affected by the hyperparameters and random seeds, leading to fragile and unpredictable results. 

\subsection{Non-Convex Optimization}
To investigate the reasons for the unsatisfactory performance of DREAMPlace and develop better methods, we look back into the problem that it wants to solve, i.e., global placement, which is a typical complex non-convex optimization problem~\cite{lu2015eplace}.~\footnote{Due to space limitation, please refer to~\cite{on_nonconvex} for more information of non-convex optimization.} 
Gradient descent converges to a first-order stationary point. When the objective function $f$ of an optimization problem is convex, this is sufficient because a first-order stationary point must be global optimal. However, when $f$ is non-convex, a first-order stationary point can be a local optimum (e.g., saddle point), and thus may be arbitrarily bad. Note that the local optima are ubiquitous in high-dimensional non-convex problems~\cite{dauphin2014identifying}. Thus, the goal of non-convex optimization is often to escape local optima and find a second-order stationary point. 

Escaping local optima and finding the global optimum is a long-standing challenge in non-convex optimization. Recent studies ~\cite{jin2017escape,egd} have shown that adding perturbation when necessary can escape local optima efficiently. That is, when the solution is close to a local optimum (i.e., the gradient has a small norm and the solution is not improved for many iterations), it will be added by a randomly sampled perturbation vector. After that, the gradient descent phase will be performed once again, and the aforementioned process will be repeated until reaching the maximum number of iterations. Since the perturbation is actually the mutation operator in the classical black-box optimization algorithm, evolutionary algorithm, these methods~\cite{jin2017escape,egd} can be seen as hybrid optimization algorithms, combining gradient-based and black-box optimization algorithms. Theoretical analysis has shown that the hybrid optimization algorithms have high probability to find an $\epsilon$-second-order stationary point by using almost the same time that gradient descent takes to find an $\epsilon$-first-order stationary point (see Theorem 4.1 in~\cite{on_nonconvex}).

\section{Method}

\subsection{Hybro Framework}
Global placement is indeed a challenging non-convex optimization problem~\cite{polar,chen2008ntuplace3,lin2020dreamplace}. In this paper, we would like to emphasize that the current leading gradient-based global placer, DREAMPlace~\cite{lin2020dreamplace,dreamplace4}, faces a number of unsatisfactory issues, mainly due to being stuck in local optima. As we mentioned before, this is a common issue for gradient-based optimizer when it is applied to solve non-convex optimization problems~\cite{on_nonconvex}. 

Inspired by escaping local optimal with perturbation~\cite{jin2017escape,egd}, we propose a hybrid optimization framework (called Hybro) for global placement, which iteratively runs a gradient-based method (i.e., DREAMPlace) and performs perturbation when it is necessary. The detailed process of Hybro is presented in Algorithm~\ref{alg:hybro}. In each iteration, Hybro uses DREAMPlace for global placement until convergence (i.e., line~3). Then, a perturbation is added to the current placement solution $\bm{s}_i$ in line~4, by using one of three strategies: Shuffle, Shuffle (all), or WireMask, which will be explained in Section~\ref{subsec-Perturb}. Hybro iterates through a predefined number of iterations, denoted as $N$. After completing $N$ iterations, it returns the best solution $\bm{s}^*$ with the smallest HPWL value, representing the optimized placement solution.

The Hybro framework combines the gradient-based iterative refinement of DREAMPlace with black-box perturbation strategies to help escape local optima, and thus improve the quality, i.e., HPWL value, of the global placement solution. After completing the global placement, one can use any tool (such as \emph{Cadence Innovus}) to proceed with the subsequent stages.

\begin{algorithm}[t!]
\caption{Hybrid Optimization (Hybro) Framework}\label{alg:hybro}
{\textbf{Parameter}:} number $N$ of iterations, perturbation strength $p\%$\\
\begin{algorithmic}[1]
\STATE Initialize the canvas, and set $i=0$;
\WHILE{$i \leq N$}
    \STATE Execute DREAMPlace for global placement to update solution $\bm{s}_i$ until it reaches convergence;
    \STATE Add a perturbation to the placement result by perturbation strategy, i.e., $\bm{s}_{i+1} \leftarrow \text{Perturb}(\bm{s}_i)$:
    \begin{itemize}
        \item Shuffle: Randomly select $p\%$ of all the macros and shuffle their locations;
        \item Shuffle (all): Randomly select $p\%$ of all the cells and shuffle their locations;
        \item WireMask: Adjust all the macros by a wire-mask-guided greedy procedure (see Algorithm 1 in~\cite{wiremask-bbo})
    \end{itemize}
    \STATE $i\leftarrow i+1$
\ENDWHILE
\STATE \textbf{return} Best solution $\bm{s}^*$ with the best HPWL value
\end{algorithmic}
\end{algorithm}

\subsection{Perturbation strategy}\label{subsec-Perturb}

Previous work has shown that an appropriate perturbation strategy is crucial for escaping local minima~\cite{jin2017escape,on_nonconvex,egd}. One basic approach is the multiple-restart gradient descent, which runs DREAMPlace independently with random initial solutions, and can be viewed as using completely random perturbation over the whole placement solution. However, it requires exponential time to find global optima~\cite{du2017gradient}, and will also be empirically shown to be inferior to our proposed methods. 

Considering the characteristics of global placement, we consider the following three perturbation methods:
\begin{itemize}
    \item Shuffle: first randomly selects a certain percentage ($p\%$) of macros and then shuffles their locations.
    \item Shuffle (all): first randomly selects a certain percentage ($p\%$) of all cells (including macros and standard cells) and then shuffles their locations.
    \item WireMask: adjusts the macros by a wire-mask-guided greedy procedure~\cite{wiremask-bbo}, which sequentially adjusts the position of each macro to the grid with the least increment of HPWL under the guidance of wire mask~\cite{lai2022maskplace}.
\end{itemize}

One key hyperparameter for Shuffle is the percentage $p\%$, which determines the perturbation strength. If $p\%$ is too high, the effect may be similar to multiple restarts, which is inefficient. If $p\%$ is too low, it may not be able to successfully escape local optima. An appropriate perturbation should have a moderate strength, helping escape the current local optima while without leaving the well-performing region. In our experiments, we find that $50\%$ is a suitable setting. Note that the WireMask perturbation strategy adjusts all macros, and thus does not have such a hyperparameter. Its robustness and superiority will be shown in our experiments. 

\section{Experiment}
\subsection{Experimental Settings}

Our framework is developed with PyTorch and CUDA, and all the experiments are conducted on Linux machines with 2.60GHz Intel Xeon CPUs, and Nvidia RTX 3090 and 3080Ti GPUs. We test different methods on the ISPD 2005~\cite{nam2005ispd2005} and ICCAD 2015 contest benchmarks~\cite{iccad15}, where the detailed statistics of these benchmarks are listed in Table~\ref{table_statistics}.

\begin{table}[htbp]
\caption{Detailed statistics of the benchmarks.}
\begin{tabular}{@{}c|ccc@{}}
\toprule
Benchmark & \#Cells   & \#Nets    & \#Pins \\ \midrule
adaptec1  & 210,904   & 221,142     & 944,053    \\
adaptec2  & 254,457   & 266,009   & 1,069,482  \\
adaptec3  & 450,927   & 466,758   & 1,875,039  \\
adaptec4  & 494,716   & 515,951   & 1,912,420  \\
bigblue1  & 277,604   & 284,479   & 1,144,691    \\
bigblue3  & 1,095,514 & 1,123,170 & 3,833,218  \\
superblue1 & 1,209,716 & 1215710 & 3,767,494\\ 
superblue3 & 1,213,253 & 1,224,979 & 3,905,321 \\
superblue4 & 795,645 & 802,513 & 2,497,940 \\
superblue5 & 1,086,888 & 1,100,825 & 3,246,878\\
superblue7 & 1,931,639 & 1,933,945 & 6,372,094 \\
superblue10 & 1,876,103 & 1,898,119 & 5,560,506\\
superblue16 & 981,559 & 999,902 & 3,013,268 \\
superblue18 & 768,068 & 771,542 & 2,559,143 \\
\bottomrule
\end{tabular}\label{table_statistics}
\end{table}

\begin{figure*}[t!]
\centering
\includegraphics[width=0.7\textwidth]{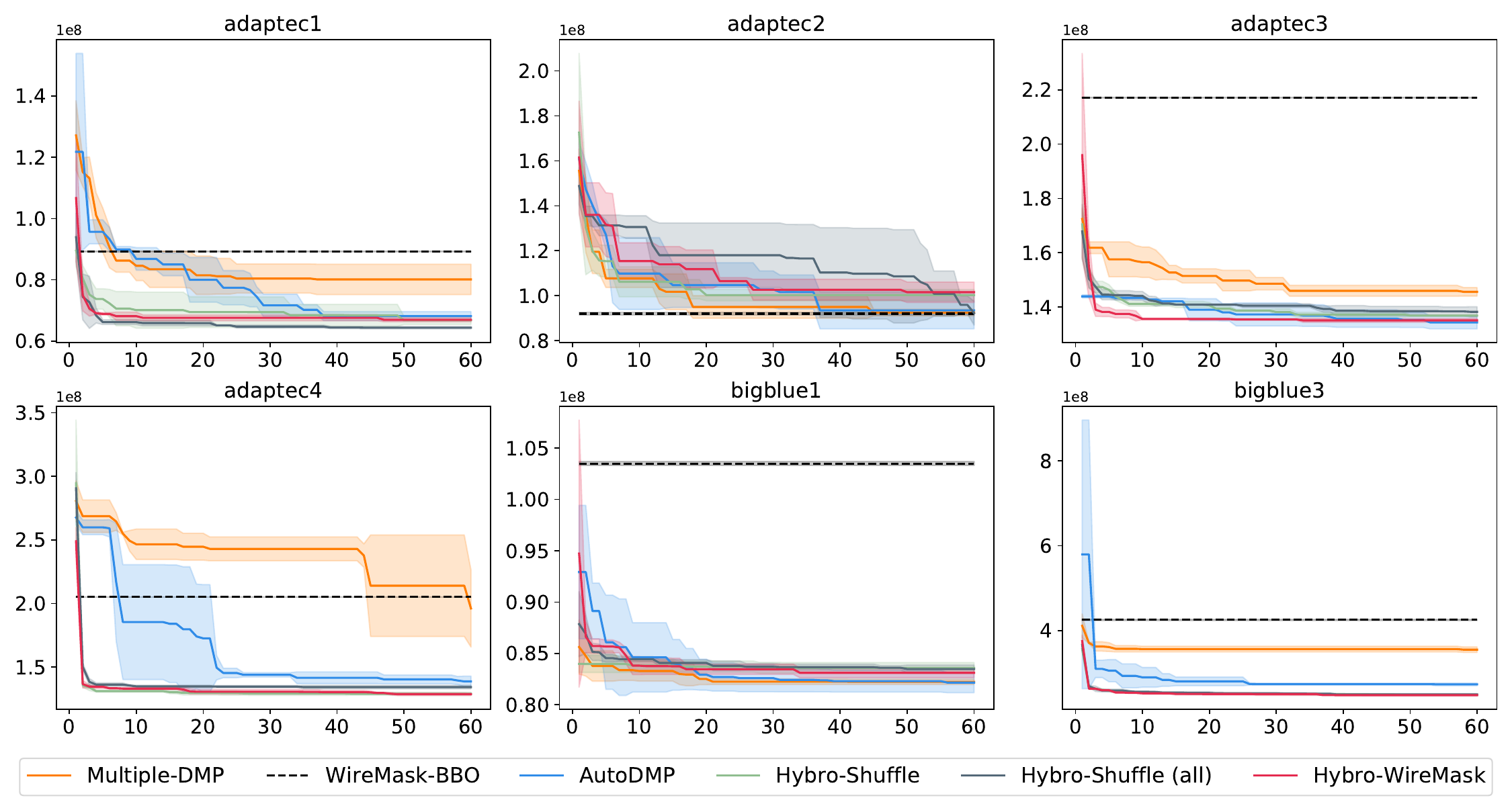}
\caption{HPWL vs. iterations of different methods on the ISPD 2005 benchmarks, where the shaded region represents the standard error derived from 3 independent runs.}
% todo change the font size of the figure
\label{fig:ispd05_hpwl}
\end{figure*}

We test the performance of the following methods for global placement:
\begin{itemize}
    \item Multiple-DMP~\cite{lin2020dreamplace,dreamplace4}: Basic baseline which runs DREAMPlace multiple times independently and returns the best found solution. For fair comparison, we use the the same number (i.e., $N$) of standard placement as Hybro.
    \item WireMask-BBO~\cite{wiremask-bbo}: A state-of-the-art black-box macro placement method. After placing the macros, it uses DREAMPlace to finish the following processes. 
    \item AutoDMP~\cite{agnesina2023autodmp}: A recently proposed global placer that improves DREAMPlace by exploring its configuration space iteratively. The number of iterations (i.e., the number of standard placement) is also set to $N$ for fair comparison. 
    \item Hybro-Shuffle: Hybro with the random shuffle strategy, where Hybro-Shuffle and Hybro-Shuffle (all) denote shuffling $p\%$ of the macros and all the cells, respectively. The perturbation strength $p\%$ is set to $50\%$.
    \item Hybro-WireMask: Hybro with the WireMask perturbation.
\end{itemize}

\subsection{ISPD 2005 Results}

Figure~\ref{fig:ispd05_hpwl} shows the curve (i.e., the change of HPWL value of the best found solution over the number of iterations) of each method on each benchmark. WireaMask-BBO solely places macros and utilizes DREAMPlace once to complete the full placement, hence resulting in a straight line. Its performance is inferior to other methods, except for adaptec2. The iterative methods all exhibit significant reduction in HPWL, which confirms that running DMP only once (corresponding to the beginning of a curve) easily leads to being stuck in local optima and is unsatisfactory. In comparison to the fully random-restart Multiple-DMP, the Hybro methods yield better placement results in most benchmarks. Moreover, in multiple benchmarks, Hybro achieves excellent HPWL values within 5 iterations, demonstrating the superiority of the Hybro framework.

\begin{table*}[t!]
% \resizebox{\columnwidth}{!}{%
\caption{HPWL values ($\times 10^7$) of full placement results. Each result consists of the mean and standard deviation of 3 runs. The best (smallest) mean value on each chip is bolded.}\label{tab:ispd05-hpwl}
\begin{tabular}{ccccccc} \toprule
    benchmark & Multiple-DMP     & WireMask-BBO             & AutoDMP                   & Hybro-Shuffle    & Hybro-Shuffle (all)      & Hybro-WireMask            \\ \midrule
    adaptec1  & 8.01 $\pm$ 0.50  & 8.93 $\pm$ 0.01          & 6.82 $\pm$ 0.15           & 6.68 $\pm$ 0.13  & \textbf{6.44 $\pm$ 0.04} & 6.70 $\pm$ 0.06           \\
    adaptec2  & 9.25 $\pm$ 0.15  & \textbf{9.20 $\pm$ 0.05} & 9.35 $\pm$ 0.82           & 10.03 $\pm$ 0.25 & 9.23 $\pm$ 0.53          & 10.15 $\pm$ 0.45          \\
    adaptec3  & 14.56 $\pm$ 0.16 & 21.72 $\pm$ 0.01         & \textbf{13.43 $\pm$ 0.24} & 13.68 $\pm$ 0.03 & 13.83 $\pm$ 0.19         & 13.51 $\pm$ 0.07          \\
    adaptec4  & 19.61 $\pm$ 3.03 & 20.51 $\pm$ 0.01         & 13.88 $\pm$ 0.42          & 12.91 $\pm$ 0.12 & 13.43 $\pm$ 0.13         & \textbf{12.87 $\pm$ 0.10} \\
    bigblue1  & 8.22 $\pm$ 0.03  & 10.35 $\pm$ 0.02         & \textbf{8.22 $\pm$ 0.10}  & 8.36 $\pm$ 0.05  & 8.35 $\pm$ 0.04          & 8.31 $\pm$ 0.05           \\
    bigblue3  & 35.49 $\pm$ 0.64 & 42.52 $\pm$ 0.11         & 27.31 $\pm$ 0.32          & 24.86 $\pm$ 0.14 & 24.93 $\pm$ 0.20         & \textbf{24.76 $\pm$ 0.07} \\ \midrule
    Avg. Rank & 4.17             & 4.83                     & 3                         & 3.17             & 2.83                     & \textbf{2.67} \\ \bottomrule
\end{tabular}%
% }
\end{table*}

Next, we show the best HPWL values achieved by the compared methods on the ISPD 2005 benchmarks. As shown in Table~\ref{tab:ispd05-hpwl}, AutoDMP and three Hybro methods are better than Multiple-DMP and WireMask-BBO. Among the three perturbation strategies of Hybro, we observe that the average rank is Hybro-WireMask < Hybro-Shuffle (all) < Hybro-Shuffle, implying that merely shuffling macros randomly may be inadequate in effectively escaping local optima on these benchmarks. Hybro-WireMask attains the best overall ranking, thereby affirming the effectiveness of WireMask as a favorable perturbation strategy.

\begin{figure}[htbp]
\centering
\includegraphics[width=0.46\textwidth]{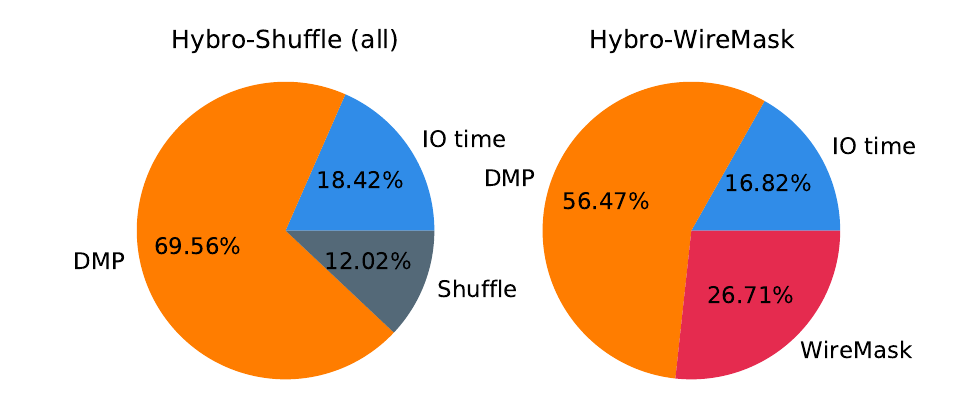}
\caption{Runtime breakdown of Hybro-Shuffle (all) and Hybro-WireMask on the ISPD 2005 benchmark \emph{adaptec3}.}
\label{fig:runtime}
\end{figure}

Additionally, we conduct a comparison of the runtime breakdown for Hybro-Shuffle (all) and Hybro-WireMask. As depicted in Figure~\ref{fig:runtime}, the runtime of each perturbation strategy occupies a small proportion of the overall time, highlighting the efficiency of our proposed perturbation strategy. WireMask requires approximately twice the amount of runtime compared to Shuffle due to the internal greedy improvement process, which is, however, still less than half of that of DMP.

\begin{figure}[t!]
\centering
\includegraphics[width=0.49\textwidth]{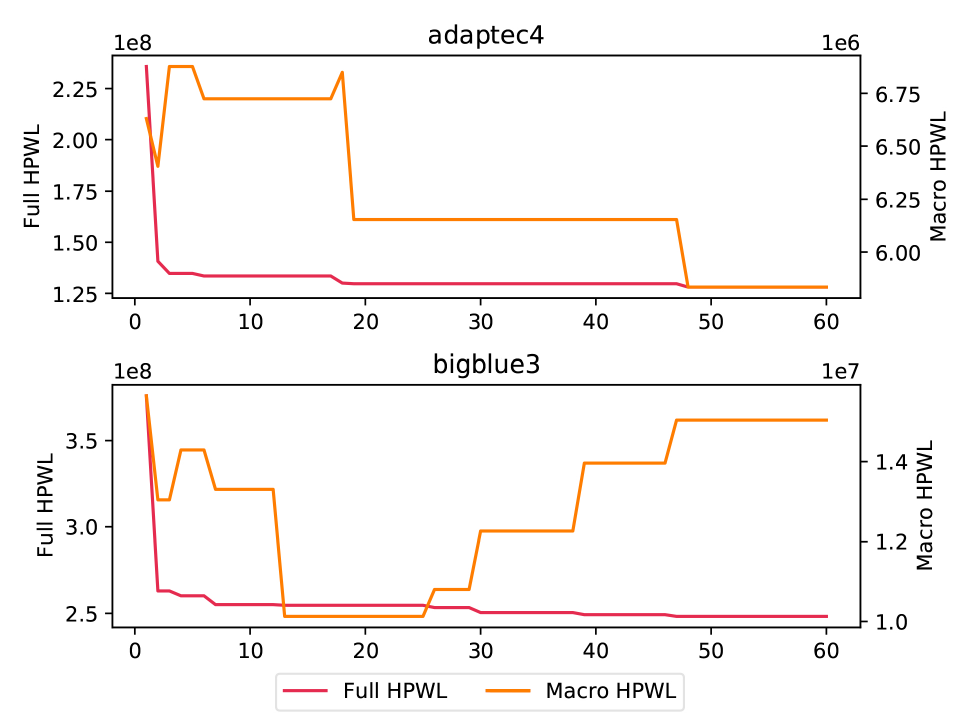}
\caption{Comparison of full placement HPWL and macro HPWL curves of Hybro-WireMask on the ISPD 2005 benchmarks \emph{adaptec4} and \emph{bigblue3}.}
\label{fig:full_macro_hpwl_curves}
\end{figure}

To examine the internal dynamics and relationship between full HPWL and macro HPWL, we plot the curves of Hybro-WireMask as shown in Figure~\ref{fig:full_macro_hpwl_curves}. The Full HPWL curves (in red) represent the HPWL value obtained from the best placement result found until the current iteration, and the macro HPWL corresponds to the HPWL value of macros for this placement result. It can be observed that there is some correlation between the macro HPWL and the full HPWL on adaptec4, while the correlation is weaker on bigblue3. However, Hybro-WireMask is capable of achieving a good full HPWL value, which ultimately aligns with the objective of global placement.

\subsection{ICCAD 2015 Results}

Next, we test the Multiple-DMP baseline and the best-performing Hybro-WireMask on the ICCAD 2015 benchmarks. After obtaining the global placement results, we use commercial tool \emph{Cadence Innovus} to evaluate their PPA metrics, including routed wirelength, routed vertical and horizontal congestion overflow, worst negative slack, total negative slack, and the number of violation points, as shown in Table~\ref{table_industrial_results}.
The results show that our Hybro-WireMask has significantly better PPA metrics than Multiple-DMP on almost all the benchmarks, except for superblue16. Although our proposed framework is not timing or congestion driven, Hybro-WireMask still achieves remarkably good timing and congestion results, demonstrating its potential. 

\begin{table*}[htbp]
\renewrobustcmd{\bfseries}{\fontseries{b}\selectfont}
\renewrobustcmd{\boldmath}{}
\newrobustcmd{\B}{\bfseries}
\caption{PPA results evaluated by \emph{Cadence Innovus} on the ICCAD 2015 benchmarks. The global placement is performed by Multiple-DMP and Hybro-WireMask, and the rest of the flow including routing and timing evaluation are performed by \emph{Cadence Innovus}. rWL is the routed wirelength; rO-V and rO-H represent the routed vertical and horizontal congestion overflow, respectively; WNS is the worst negative slack; TNS is the total negative slack; NVP is the number of violation points. WNS and TNS are the larger the better, while the other metrics are the smaller the better. The best result of each metric on each benchmark is bolded.}
\begin{tabular}{@{}c|cccccc|cccccc@{}}
\toprule
\multirow{3}{*}{Benchmark}         &\multicolumn{6}{c}{Multiple-DMP}    & \multicolumn{6}{c}{Hybro-WireMask} \\
            & rWL  &  rO-V & rO-H & WNS & TNS & NVP  & rWL  &  rO-V & rO-H & WNS & TNS & NVP \\
            & (m)  &  (\%) & (\%) & (ns) & ($\upmu$s) & \#  & (m)  &  (\%) & (\%) & (ns) & ($\upmu$s) & \# \\
\midrule
superblue1  & 108.78 & 1.00 & 0.32 & -40.715 & -107.32 & 18721       & \B 92.30  & \B 0.09 & \B 0.04 & \B -32.114 & \B -51.79  & \B 10452 \\
superblue3  & 123.41 & 1.17 & 0.25 & \B -46.010 & -32.98  & 7968        & \B 100.62 & \B 0.23 & \B 0.04 & -47.259 & \B -19.23  & \B 6316  \\
superblue4  & 74.17  & 0.37 & 0.15 & \B -32.717 & -49.91  & 9212        & \B 69.21  & \B 0.19 & \B 0.09 & -34.129 & \B -41.46  & \B 7139  \\
superblue5  & 134.09 & 2.32 & 0.51 & \B -54.052 & -49.23  & 11677       & \B 112.40 & \B 1.31 & \B 0.11 & -54.310 & \B -38.04  & \B 9889  \\
superblue7  & 209.11 & 5.10 & 4.18 & -40.175 & -113.31 & 34744       & \B 132.00 & \B 0.04 & \B 0.03 & \B -16.887 & \B -21.84  & \B 11216 \\
superblue10 & 207.17 & 1.38 & 0.63 & -72.977 & -174.62 & 18465       & \B 165.91 & \B 0.14 & \B 0.10 & \B -40.209 & \B -104.42 & \B 14691 \\
superblue16 & \B 95.08  & \B 1.16 & \B 0.28 & \B -32.729 & \B -72.64  & \B 19354       & 112.16 & 5.91 & 0.74 & -53.859 & -102.12 & 25685 \\
superblue18 & 51.46  & \B 0.10 & 0.02 & -17.034 & -19.68  & 5570        & \B 48.93  & 0.17 & \B 0.01 & \B -15.752 & \B -11.76  & \B 5001  \\
\bottomrule
\end{tabular}\label{table_industrial_results}
\end{table*}

\begin{figure*}[t!]
\centering
\begin{minipage}{0.24\textwidth}
\centering
\includegraphics[width=\textwidth, height=4cm, keepaspectratio]{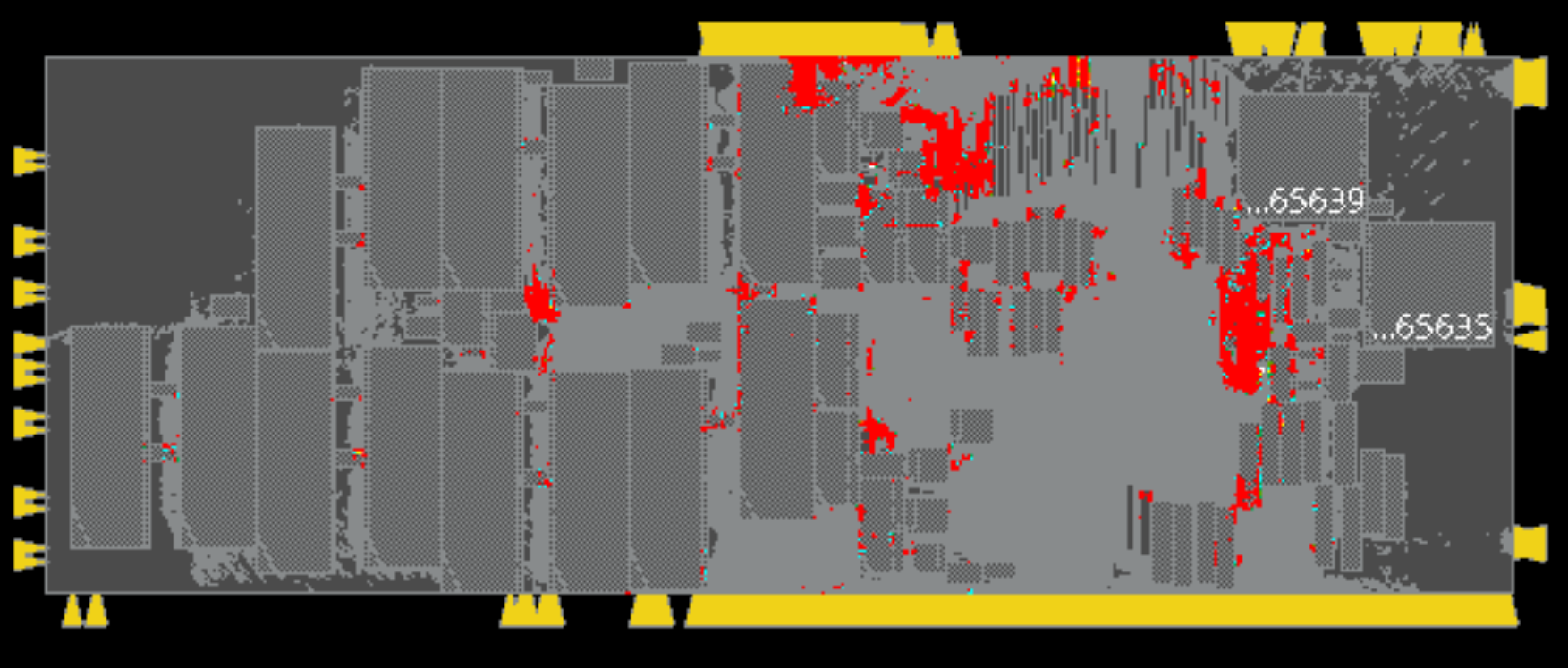}
\end{minipage}
\begin{minipage}{0.24\textwidth}
\centering
\includegraphics[width=\textwidth, height=4.8cm, keepaspectratio]{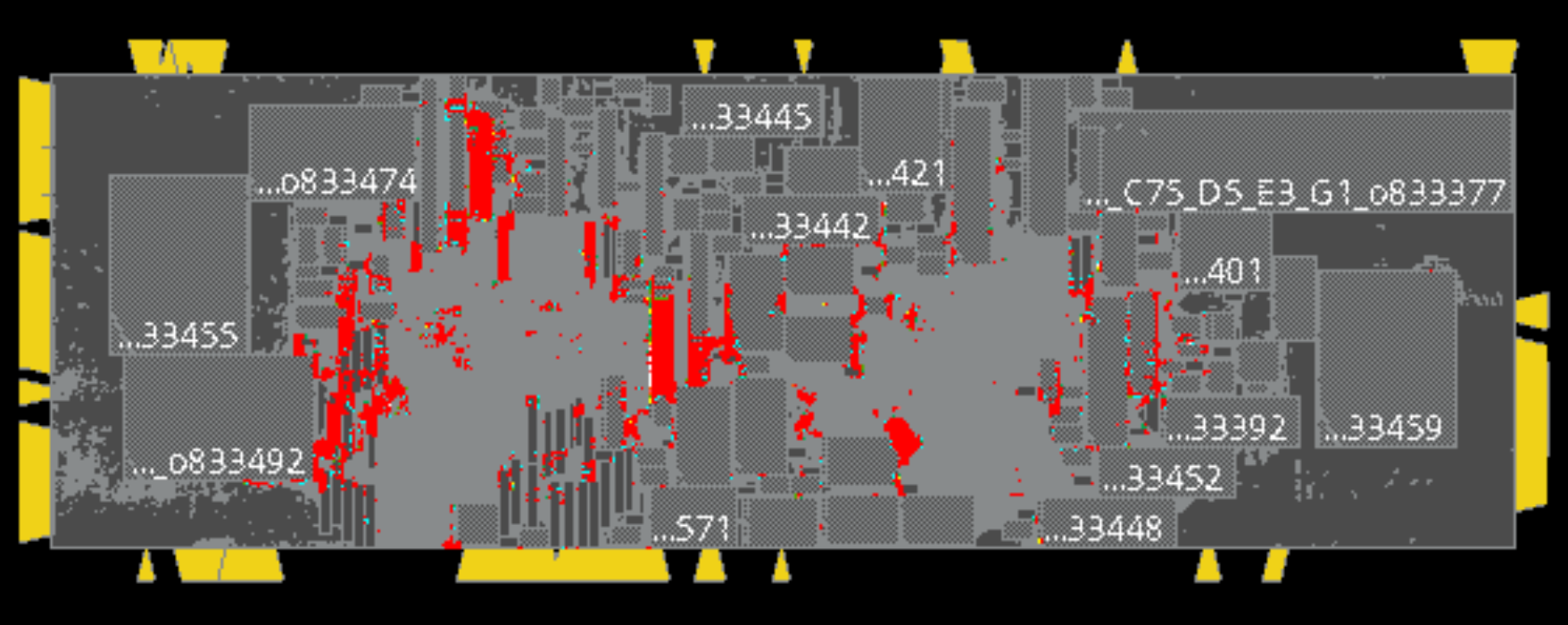}
\end{minipage}
\begin{minipage}{0.24\textwidth}
\centering
\includegraphics[width=\textwidth, height=4cm, keepaspectratio]{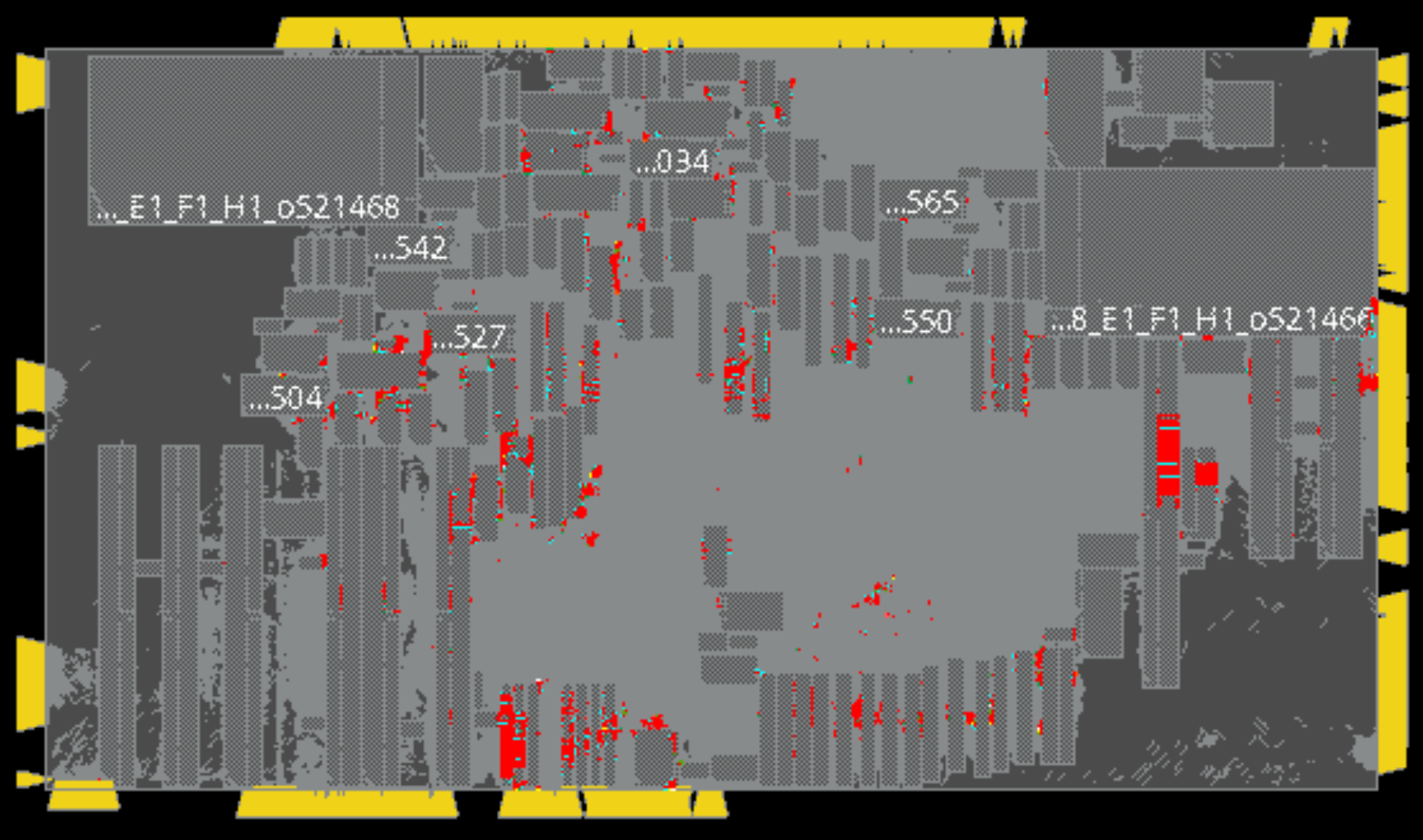}
\end{minipage}
\begin{minipage}{0.24\textwidth}
\centering
\includegraphics[width=\textwidth, height=3cm, keepaspectratio]{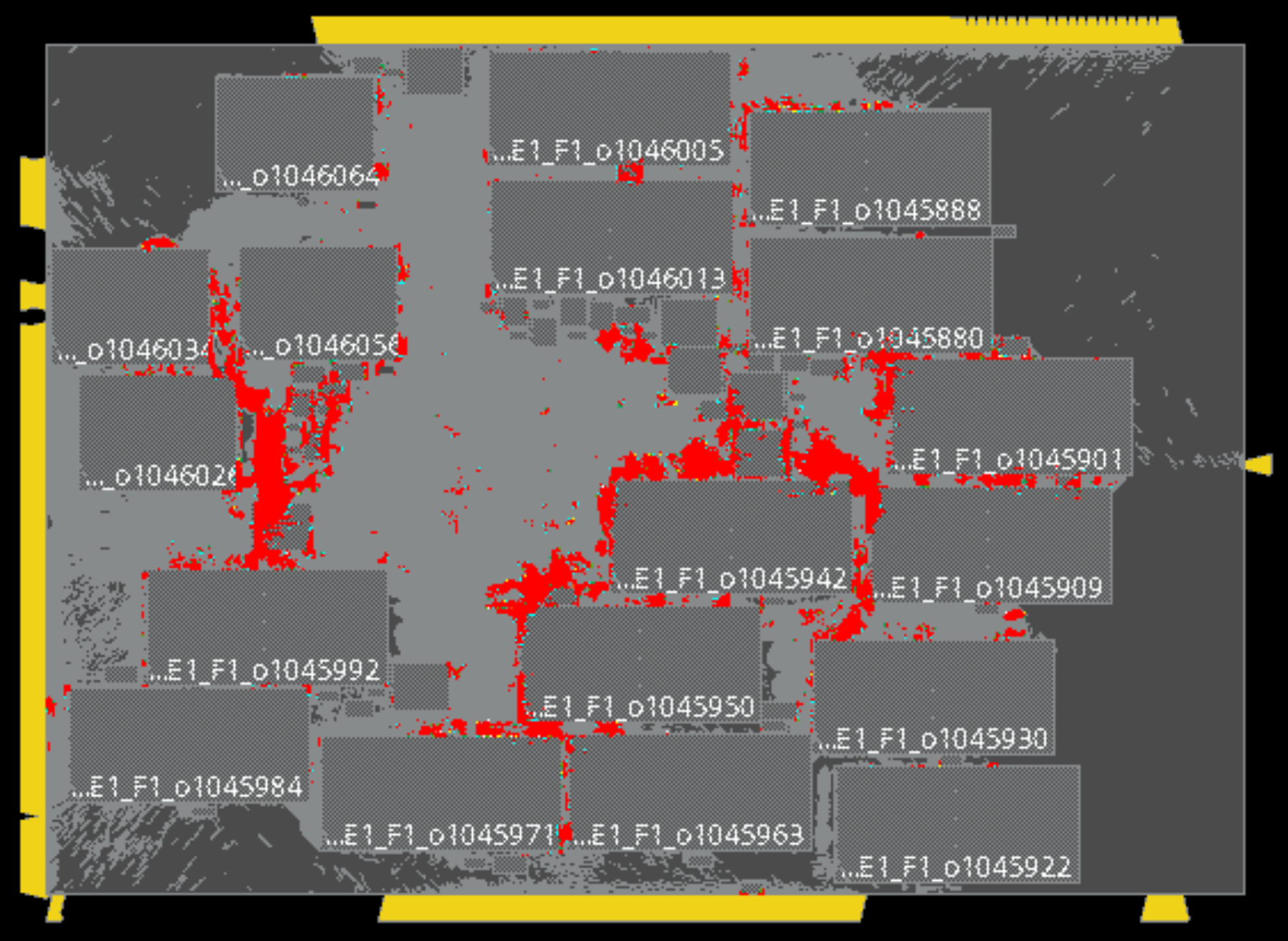}
\end{minipage} \
\begin{minipage}{0.24\textwidth}
\centering
\includegraphics[width=\textwidth, height=4cm, keepaspectratio]{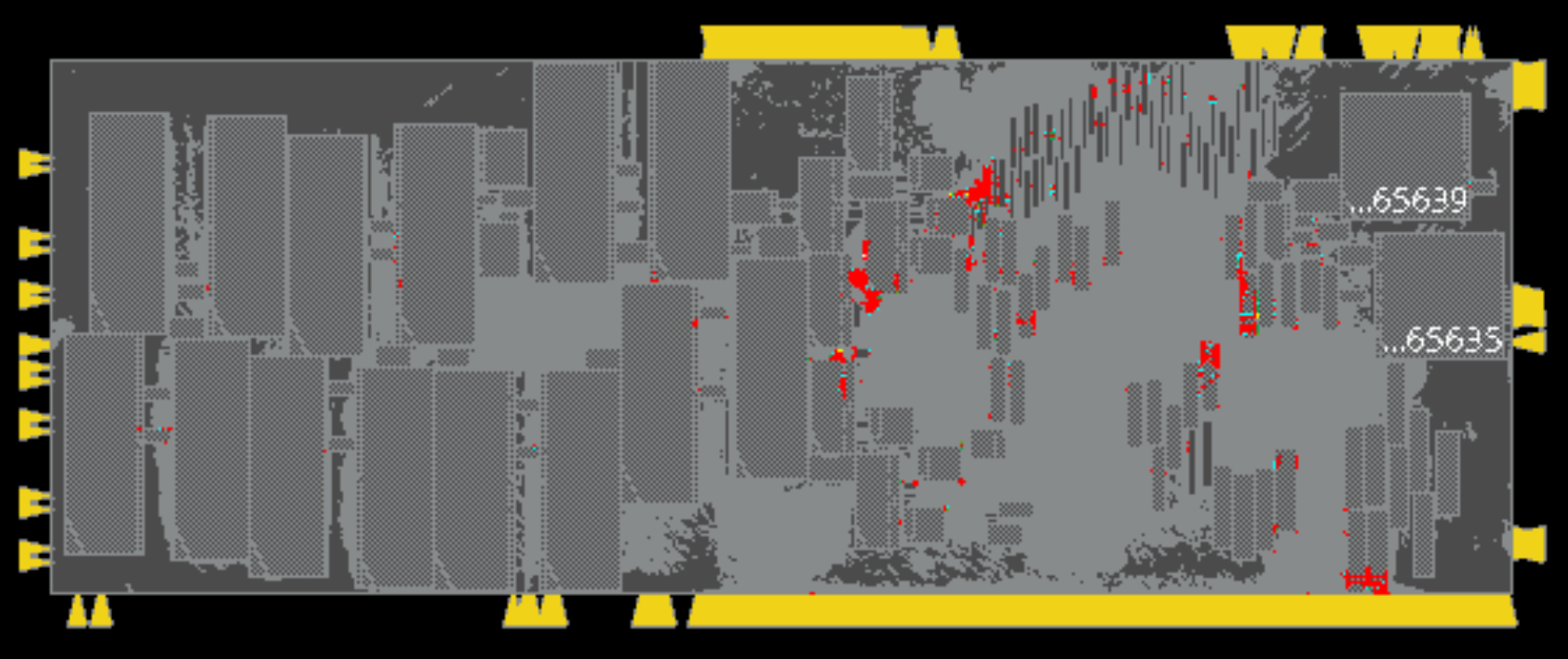}
\end{minipage}
\begin{minipage}{0.24\textwidth}
\centering
\includegraphics[width=\textwidth, height=4.8cm, keepaspectratio]{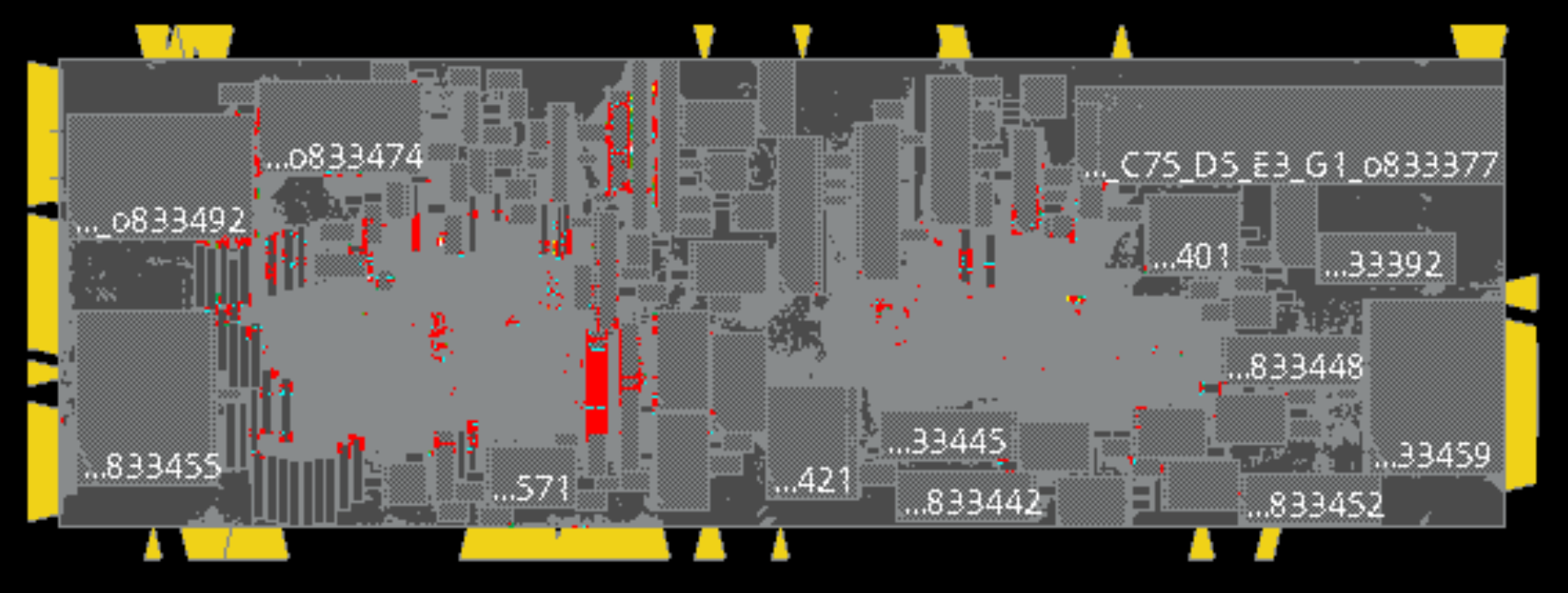}
\end{minipage}
\begin{minipage}{0.24\textwidth}
\centering
\includegraphics[width=\textwidth, height=4cm, keepaspectratio]{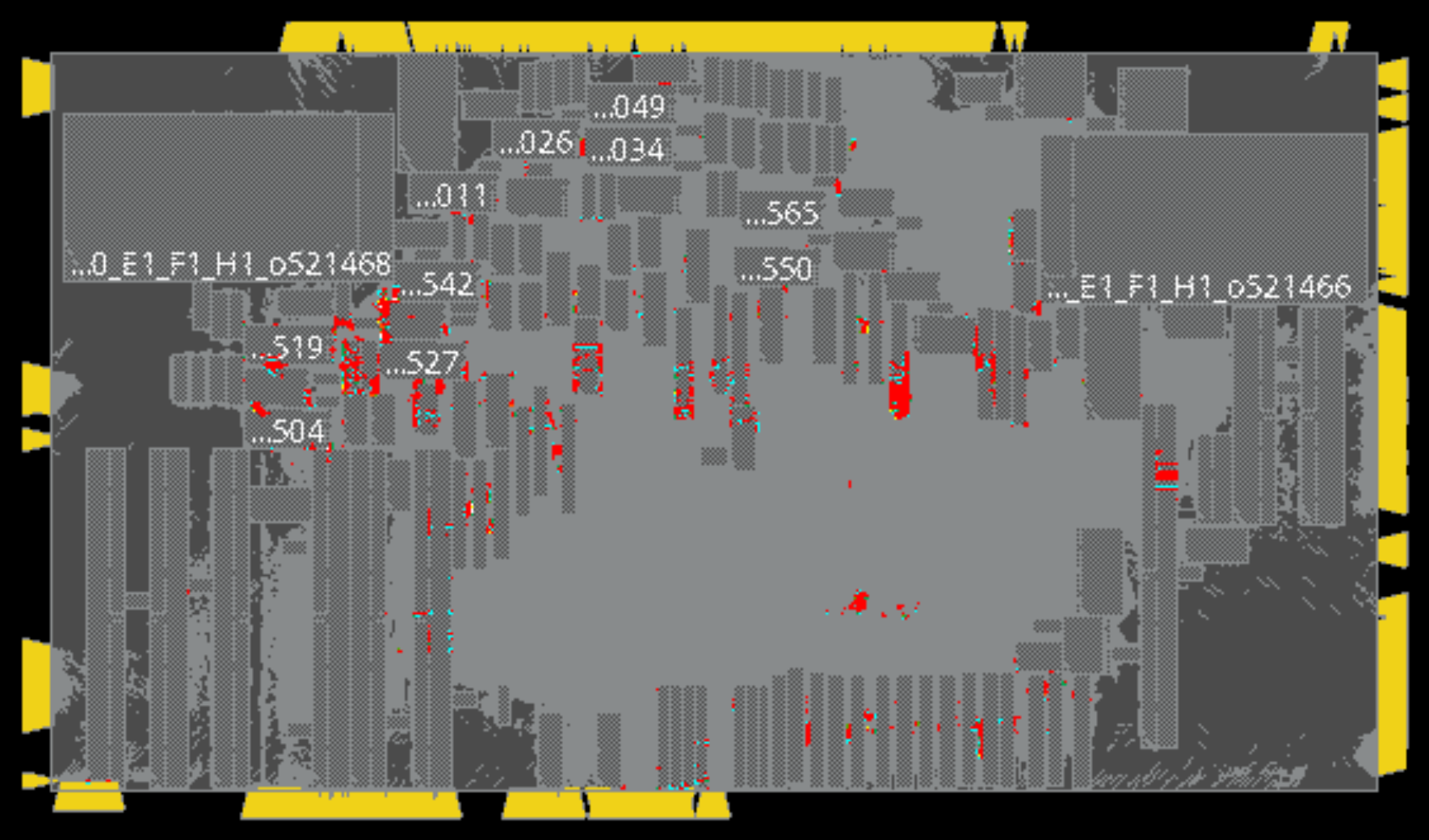}
\end{minipage}
\begin{minipage}{0.24\textwidth}
\centering
\includegraphics[width=\textwidth, height=3cm, keepaspectratio]{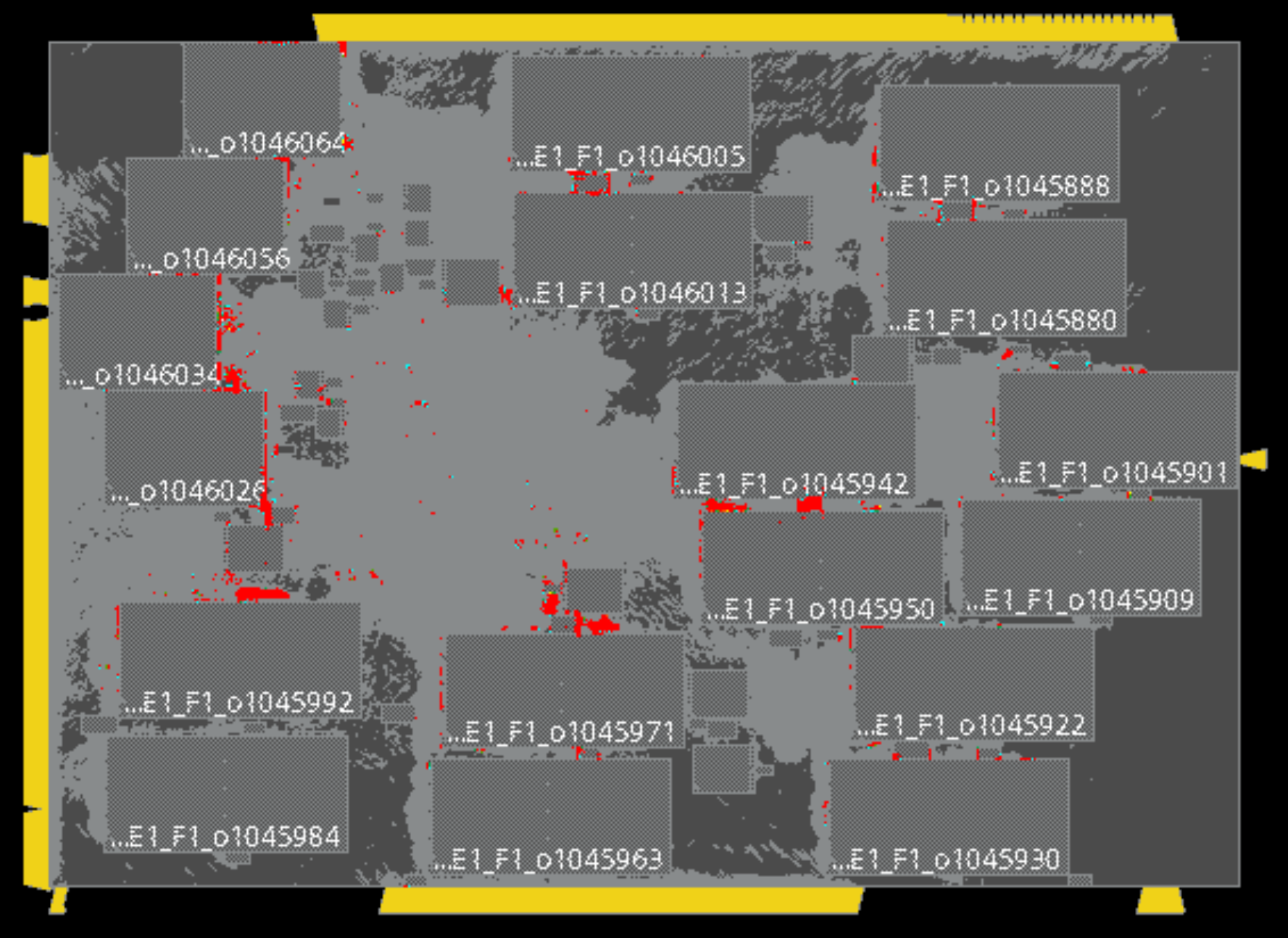}
\end{minipage} \
\begin{minipage}{0.24\textwidth}
\centering
\textbf{(a) superblue1}
\end{minipage}
\begin{minipage}{0.24\textwidth}
\centering
\textbf{(b) superblue3}
\end{minipage}
\begin{minipage}{0.24\textwidth}
\centering
\textbf{(c) superblue4}
\end{minipage}
\begin{minipage}{0.24\textwidth}
\centering
\textbf{(d) superblue10}
\end{minipage}
\caption{Placement layouts and congestions of Multiple-DMP (top row) and Hybro-WireMask (bottom row) on the ICCAD 2015 benchmarks, superblue1, superblue3, superblue4, and superblue10. The congestion results are obtained by \emph{Cadence Innovus}, where red points indicate the congestion critical regions.}
\label{fig:congestion}
\end{figure*}

Finally, we compare the placement layouts and congestions (red points) of Multiple-DMP and Hybro-WireMask on the ICCAD 2015 benchmarks, superblue1, superblue3, superblue4, and superblue10, as shown in Figure~\ref{fig:congestion}. The congestion results are obtained by \emph{Cadence Innovus}. It is evident that the placement layouts obtained by our algorithm exhibit superiority.

\section{Conclusion}
In this work, we propose Hybro, a hybrid optimization framework that can help global placer in escaping local optima. Experimental results on the ISPD 2005 and ICCAD 2015 benchmarks show that Hybro is efficient yet effective. Future works include applying Hybro to other analytical global placers and timing-driven placement. 

\printbibliography

@inproceedings{nam2005ispd2005,
  title={The {ISPD}2005 placement contest and benchmark suite},
  author={Nam, Gi-Joon and Alpert, Charles J and Villarrubia, Paul and Winter, Bruce and Yildiz, Mehmet},
  booktitle={Proceedings of the 9th International Symposium on Physical Design},
  pages={216--220},
  year={2005},
  address = {San Francisco, CA}
}

@article{lu2015eplace,
  title={e{P}lace: {E}lectrostatics-based placement using fast {F}ourier transform and {N}esterov's method},
  author={Lu, Jingwei and Chen, Pengwen and Chang, Chin-Chih and Sha, Lu and Huang, Dennis Jen-Hsin and Teng, Chin-Chi and Cheng, Chung-Kuan},
  journal={ACM Transactions on Design Automation of Electronic Systems},
  volume={20},
  number={2},
  pages={1--34},
  year={2015},
}

@article{cheng2018replace,
  title={{R}eplace: {A}dvancing solution quality and routability validation in global placement},
  author={Cheng, Chung-Kuan and Kahng, Andrew B and Kang, Ilgweon and Wang, Lutong},
  journal={IEEE Transactions on Computer-Aided Design of Integrated Circuits and Systems},
  volume={38},
  number={9},
  pages={1717--1730},
  year={2018},
}

@article{lin2020dreamplace,
  title={{DREAMP}lace: {D}eep learning toolkit-enabled gpu acceleration for modern {VLSI} placement},
  author={Lin, Yibo and Jiang, Zixuan and Gu, Jiaqi and Li, Wuxi and Dhar, Shounak and Ren, Haoxing and Khailany, Brucek and Pan, David Z},
  journal={IEEE Transactions on Computer-Aided Design of Integrated Circuits and Systems},
  volume={40},
  number={4},
  pages={748--761},
  year={2020},
}

@inproceedings{agnesina2023autodmp,
  title={Auto{DMP}: {A}utomated {DREAMP}lace-based macro placement},
  author={Agnesina, Anthony and Rajvanshi, Puranjay and Yang, Tian and Pradipta, Geraldo and Jiao, Austin and Keller, Ben and Khailany, Brucek and Ren, Haoxing},
  booktitle={Proceedings of the 27th International Symposium on Physical Design},
  pages={149--157},
  year={2023},
  address={Virtual}
}

@article{murata1996vlsi,
  title={{VLSI} module placement based on rectangle-packing by the sequence-pair},
  author={Murata, Hiroshi and Fujiyoshi, Kunihiro and Nakatake, Shigetoshi and Kajitani, Yoji},
  journal={IEEE Transactions on Computer-Aided Design of Integrated Circuits and Systems},
  volume={15},
  number={12},
  pages={1518--1524},
  year={1996},
  publisher={IEEE}
}

@inproceedings{chang2000b,
  title={B*-trees: {A} new representation for non-slicing floorplans},
  author={Chang, Yun-Chih and Chang, Yao-Wen and Wu, Guang-Ming and Wu, Shu-Wei},
  booktitle={Proceedings of the 37th Annual Design Automation Conference},
  pages={458--463},
  year={2000},
  address={Los Angeles, CA}
}

@article{chen2008ntuplace3,
  title={NTUplace3: {A}n analytical placer for large-scale mixed-size designs with preplaced blocks and density constraints},
  author={Chen, Tung-Chieh and Jiang, Zhe-Wei and Hsu, Tien-Chang and Chen, Hsin-Chen and Chang, Yao-Wen},
  journal={IEEE Transactions on Computer-Aided Design of Integrated Circuits and Systems},
  volume={27},
  number={7},
  pages={1228--1240},
  year={2008},
  publisher={IEEE}
}

@inproceedings{
  lai2022maskplace,
  title={MaskPlace: {F}ast chip placement via reinforced visual representation learning},
  author={Yao Lai and Yao Mu and Ping Luo},
  booktitle={Advances in Neural Information Processing Systems 35},
  year={2022},
  address={New Orleans, LA}
}

@inproceedings{hsu2011tsv,
  title={{TSV}-aware analytical placement for 3{D} {IC} designs},
  author={Hsu, Meng-Kai and Chang, Yao-Wen and Balabanov, Valeriy},
  booktitle={Proceedings of the 48th Design Automation Conference},
  pages={664--669},
  year={2011},
  address={San Diego, CA}
}

@inproceedings{jin2017escape,
  title={How to escape saddle points efficiently},
  author       = {Chi Jin and
                  Rong Ge and
                  Praneeth Netrapalli and
                  Sham M. Kakade and
                  Michael I. Jordan},
  booktitle={Proceedings of the 34th International Conference on Machine Learning},
  pages={1724--1732},
  year={2017},
  address={Sydney, Australia}
}

@inproceedings{egd,
  author       = {Ke Xue and
                  Chao Qian and
                  Ling Xu and
                  Xudong Fei},
  title        = {Evolutionary Gradient Descent for Non-convex Optimization},
  booktitle    = {Proceedings of the Thirtieth International Joint Conference on Artificial
                  Intelligence},
  pages        = {3221--3227},
  year         = {2021},
  address = {Montreal, Canada}
}

@inproceedings{du2017gradient,
  title={Gradient descent can take exponential time to escape saddle points},
  author       = {Simon S. Du and
                  Chi Jin and
                  Jason D. Lee and
                  Michael I. Jordan and
                  Aarti Singh and
                  Barnab{\'{a}}s P{\'{o}}czos},
  booktitle={Advances in Neural Information Processing Systems 30},
  pages={1067--1077},
  year={2017},
  address={Long Beach, CA}
}

@article{dreamplace4,
  author       = {Peiyu Liao and
                  Dawei Guo and
                  Zizheng Guo and
                  Siting Liu and
                  Yibo Lin and
                  Bei Yu},
  title        = {DREAMPlace 4.0: {T}iming-Driven Placement With Momentum-Based Net Weighting
                  and Lagrangian-Based Refinement},
  journal={IEEE Transactions on Computer-Aided Design of Integrated Circuits and Systems}, 
  volume       = {42},
  number       = {10},
  pages        = {3374--3387},
  year         = {2023}
}

@inproceedings{
wiremask-bbo,
title={Macro Placement by Wire-Mask-Guided Black-Box Optimization},
author={Yunqi Shi and Ke Xue and Lei Song and Chao Qian},
booktitle={Advances in Neural Information Processing Systems 36},
year={2023},
address={New Orleans, LA}
}

@inproceedings{iccad15,
  author       = {Myung{-}Chul Kim and
                  Jin Hu and
                  Jiajia Li and
                  Natarajan Viswanathan},
  title        = {{ICCAD-2015} {CAD} Contest in Incremental Timing-driven Placement
                  and Benchmark Suite},
  booktitle    = {Proceedings of the {IEEE/ACM} International Conference on Computer-Aided
                  Design},
  pages        = {921--926},
address={Austin, TX},
  year         = {2015},
}

@inproceedings{dauphin2014identifying,
  title={Identifying and attacking the saddle point problem in high-dimensional non-convex optimization},
  author       = {Yann N. Dauphin and
                  Razvan Pascanu and
                  {\c{C}}aglar G{\"{u}}l{\c{c}}ehre and
                  KyungHyun Cho and
                  Surya Ganguli and
                  Yoshua Bengio},
  booktitle={Advances in Neural Information Processing Systems 27},
  pages={2933--2941},
  year={2014},
  address={Montr{\'{e}}al, Canada}
}

@article{on_nonconvex,
author = {Jin, Chi and Netrapalli, Praneeth and Ge, Rong and Kakade, Sham M. and Jordan, Michael I.},
title = {On Nonconvex Optimization for Machine Learning: {G}radients, Stochasticity, and Saddle Points},
year = {2021},
publisher = {Association for Computing Machinery},
address = {New York, NY, USA},
volume = {68},
number = {2},
journal = {Journal of the ACM},
}

@article{essential-issues-in-analytical,
  author       = {Yao{-}Wen Chang and
                  Zhe{-}Wei Jiang and
                  Tung{-}Chieh Chen},
  title        = {Essential Issues in Analytical Placement Algorithms},
  journal={IPSJ Transactions on System LSI Design Methodology},
  volume={2},
  pages={145--166},
  year={2009},
}

@article{ripple,
  author       = {Xu He and
                  Tao Huang and
                  Linfu Xiao and
                  Haitong Tian and
                  Evangeline F. Y. Young},
  title        = {Ripple: {A} Robust and Effective Routability-Driven Placer},
  journal      = {{IEEE} Transactions on Computer-Aided Design of Integrated Circuits and Systems},
  volume       = {32},
  number       = {10},
  pages        = {1546--1556},
  year         = {2013},
}

@inproceedings{polar,
  author       = {Tao Lin and
                  Chris C. N. Chu and
                  Gang Wu},
  title        = {{POLAR} 3.0: {A}n Ultrafast Global Placement Engine},
  booktitle    = {Proceedings of the {IEEE/ACM} International Conference on Computer-Aided
                  Design},
  pages        = {520--527},
  address = {Austin, TX},
  year         = {2015},
}
% todo 引文控制在30篇以内就好，不能太多了。

\end{document}